\pdfoutput=1
\documentclass[11pt]{article}
\usepackage{amsmath, bm}

\usepackage[preprint]{acl}
\usepackage{changepage,threeparttable}
\usepackage{times}
\usepackage{latexsym}
\usepackage{longtable}
\usepackage{array}
\usepackage{float}
\usepackage{booktabs, multirow}
\usepackage{lscape}
\usepackage{longtable}
\usepackage[normalem]{ulem}
\useunder{\uline}{\ul}{}
\usepackage[toc,page]{appendix}

\usepackage[T1]{fontenc}
\usepackage[utf8]{inputenc}

\usepackage{caption}
\usepackage{placeins} % In the preamble

\usepackage{makecell}
\usepackage{afterpage}

\usepackage{microtype}

\usepackage{inconsolata}

\usepackage{graphicx}
\usepackage{tabularx}
\usepackage{geometry}
\usepackage{array}
\usepackage{lipsum} % For dummy text
\usepackage{xcolor,colortbl}

\title{From Measurement Instruments to Data: Leveraging Theory-Driven Synthetic Training Data for Classifying Social Constructs}

\author{Lukas Birkenmaier \\ GESIS, Germany \\  
RWTH - Aachen University \\ \texttt{lukas.birkenmaier@gesis.org}
         \And  Matthias Roth \\ GESIS, Germany \\ \texttt{matthias.roth@gesis.org} 
         \AND 
         Indira Sen \\ University of Mannheim, Germany \\ \texttt{indira.sen@uni-mannheim.de}}

\begin{document}
\maketitle
\begin{abstract}
Computational text classification is a challenging task, especially for multi-dimensional social constructs. Recently, there has been increasing discussion that synthetic training data could enhance classification by offering examples of how these constructs are represented in texts. 
In this paper, we systematically examine the potential of \textit{theory-driven} synthetic training data for improving the measurement of social constructs. In particular, we explore how researchers can transfer established knowledge from measurement instruments in the social sciences, such as survey scales or annotation codebooks, into theory-driven generation of synthetic data. Using two studies on measuring sexism and political topics, we assess the added value of synthetic training data for fine-tuning text classification models. 
Although the results of the sexism study were less promising, our findings demonstrate that synthetic data can be highly effective in reducing the need for labeled data in political topic classification. With only a minimal drop in performance, synthetic data allows for substituting large amounts of labeled data. Furthermore, theory-driven synthetic data performed markedly better than data generated without conceptual information in mind.

\end{abstract}

\section{Introduction}

High-quality labeled data are essential for text classification models to be useful research tools for the computational social sciences, both for training and fine-tuning machine learning models, but also for evaluating them. 
However, high-quality training data is often scarce. On the one hand, this can be due to factors that affect the labeling process, such as lack of expertise and dedication from coders \citep{zhang2016learning} or coding instructions that fail to incorporate subjectivity and disagreement into the labeling process \citep{plank2022problem, uma2021learning}. On the other hand, external factors also affect the availability of labeled data, such as financial or organizational resources, privacy concerns or copyright issues. 
To tackle this challenge, recent studies explore the potential of generative models to create synthetic data for training and fine-tuning machine learning models \citep{li2023synthetic,guo2024generativeaisyntheticdata,achman}. Synthetic data created by generative Large-Language Models (LLMs) can be used to augment existing datasets or create entirely new datasets in cases where no natural training data exists. While synthetic training data is gaining traction to enhance the conceptual range and richness of datasets, it is not clear how to best control the generation of synthetic training data by LLMs for common use cases in the computational social sciences.

Thus, we propose the need for a structured evaluation of synthetic training data generation strategies to improve the measurement of social constructs. In this study, we explore the potential of \textit{theory-driven} synthetic training data. We define theory-driven synthetic data as data that was generated based on established social science measurement instruments or any other detailed conceptual framework related to social science constructs. We conduct two studies focusing on different constructs—political topics and sexism—using distinct measurement instruments, namely annotation codebooks and survey scales. Our primary question is: \textbf{Can theory-driven synthetic training data improve the measurement of social constructs?} To answer this question, we orient along the following assumptions:

\begin{enumerate} 
\itemsep0em
\item Incorporating theory-driven synthetic training data \textit{improves the overall performance} for in-domain and out-of-domain (OOD) classification \item Theory-driven synthetic training data can partially \textit{substitute human-labeled data} while maintaining comparable performance for in-domain and OOD classification

\end{enumerate}

The results of our study provide an ambivalent picture. For the political topics study utilizing annotation codebooks, we see only minimal drops or even slight performance increases when substituting 30-90\% of real data with synthetic data, indicating that synthetic data can help reduce annotation costs for training political topic classifiers. 
By contrast, in the sexism study utilizing survey scales, incorporating theory-driven synthetic data did not yield performance improvements, and models trained on varying proportions of synthetic data performed significantly worse for larger shares of synthetic data.
We discuss possible determinants and implications of these findings, and release our code and data to facilitate future research here (anonymized). 

\section{Related Work}

\subsection{LLMs as a source of synthetic training data}

LLMs have been used to create synthetic data since their development. At first, smaller models such as GPT-2 were fine-tuned to generate training examples for specific classification tasks \cite{juuti2020littlegoeslongway,wullach-etal-2021-fight-fire,schmidhuber-2021-universitat}. However because fine-tuning state-of-the-art LLMs became more cost-intensive and access to LLMs shifted to APIs, researchers started to explore prompt-based approaches of synthetic data creation \cite{han_ptr_2022}, which rely on few-shot learning capabilities of LLMs \cite{brown2020languagemodelsfewshotlearners}. Under the prompt-based approach, the primary way to steer the data-generating process of the LLM is to add conditioning information to the prompt. Previous work can broadly be categorized by their prompting strategies: (a) whether they add theoretical information on the construct of interest to the prompt, and (b) whether they provide examples of existing data to generate alternations (see Table \ref{tab:overview})

\begin{table}
    \centering
    \resizebox{\columnwidth}{!}{%
\begin{tabular}{llcc}
\hline
\textbf{Studies}                                                  & \textbf{Construct(s)}                                                 & \textbf{Alternations} & \textbf{\begin{tabular}[c]{@{}c@{}}Theory\\ -Driven\end{tabular}} \\ \hline
\citeauthor{achman}                                  & \begin{tabular}[c]{@{}l@{}}Political calls\\ to action\end{tabular}   & Yes                   & No                                                                \\
\citeauthor{Ghanadian_2024}                         & \begin{tabular}[c]{@{}l@{}}Suicidal \\ ideation\end{tabular}          & -                     & Yes                                                               \\
\citeauthor{hui2024toxicraftnovelframeworksynthetic} & \begin{tabular}[c]{@{}l@{}}Harmful \\ infomation\end{tabular}         & -                     & No                                                                \\
\citeauthor{li2023synthetic}                         & Multiple (10)                                                         & Yes                   & No                                                                \\
\citeauthor{mahmoudi2024zero}                        & \begin{tabular}[c]{@{}l@{}}Stance \\ detection\end{tabular}           & Yes                   & No                                                                \\
\citeauthor{moller2023parrot}                        & Multiple (10)                                                         & Yes                   & No                                                                \\
\citeauthor{veselovsky2023generating}                & Sarcasm                                                               & Yes                   & No                                                                \\
\citeauthor{wagner2024power}                         & \begin{tabular}[c]{@{}l@{}}Stance \\ detection\end{tabular}           & No                    & No                                                                \\ \hline
This study                                                        & \begin{tabular}[c]{@{}l@{}}Sexism and\\ Political Topics\end{tabular} & Yes                   & Yes                                                               \\ \hline
\end{tabular}
    }
    \caption{\textbf{Studies on synthetic training data generation.} We differentiate our work from previous research based on whether prompts ask the LLM to rewrite examples from real data (alterations) and whether they include additional theoretical information on the construct to be classified. "-" indicates that the exact prompt could not be determined from the information provided by the authors. See Appendix \ref{exp} for a more detailed overview.}
    \label{tab:overview}
\end{table}

Almost all studies shown in Table \ref{tab:overview} use examples randomly drawn from a real-world dataset in their prompt. If they do so, the also instruct the LLM to base the newly generated training examples on the examples provided in the prompt. \citealp{li2023synthetic} and \citealp{veselovsky2023generating} additionally compared the presence of an example in a prompt to its absence and concluded that providing examples in the prompt increases the performance in downstream classifiers.

Likewise, some studies add theoretical information on the construct of interest to the prompts. In their study on sarcasm detection, \citealp{veselovsky2023generating} investigated the inclusion of a taxonomy of sarcasm in their prompts, although the taxonomy does not draw on a social science theory. They conclude that including the taxonomy of sarcasm did not increase downstream classifier performance. In contrast, their study on suicidal ideation detection \citealp{Ghanadian_2024} found that including information which was drawn from a review of relevant psychological literature improved the performance of the downstream classifier.

%Given that only a few studies systematically investigated the addition of theoretical information to the prompt and the inconclusive results of these studies, this study aims to contribute to the research by systematically evaluating the effect of introducing theoretical information. 

\subsection{The Potential of Theory-Driven Synthetic Data}

We investigate two sources of information originating from social science measurement instruments to generate theory-driven synthetic training data: survey questions and annotation codebooks. Both survey questions and codebooks are readily available sources of conceptual information in the social sciences.

Survey questions aim to measure the theoretically grounded attributes of respondents. The process of questionnaire design involves multiple checks of their validity, such as cognitive pre-testing \cite{groves_survey_2011} or the analysis of their psychometric properties \cite{furr_psychometrics_2008, repke2024validity}. Thus, survey questions present high-quality text data \cite{fang2022evaluating}, which can be used in prompts to guide the generation process of LLMs. 

Annotation codebooks are developed to annotate existing text data with labels. Like survey questions, codebooks are created in a theory-driven process, providing a rationale for how social science constructs, such as populism or sexism, should manifest in text data. They are particularly well suited for creating multi-label synthetic training data, as they provide information on many aspects of a theoretical construct.

Subsequently, we outline our research design to test the effectiveness of theory-driven synthetic data based on these measurement instruments. 

\section{Method}

\subsection{Data and Case Selection}

We rely on a two-class (sexism) and a multi-class (political topics) dataset to explore the effect of theory-driven synthetic data for training text classification models. Both studies serve as important case studies because they focus on constructs commonly explored by computational social scientists, who seek to address social science questions through natural language processing methods \citep{glavas-etal-2019-computational}. We include both in-domain datasets that are used to train and test the models, as well as OOD datasets used for evaluation only. Table \ref{table:political_topics} and Table \ref{table:sexism_study} provide an overview of the class distribution for each dataset.

\begin{table*}[h!]
\centering
\small
\begin{tabular}{llccccccc}
\toprule
\multicolumn{9}{c}{\textbf{Political Topics Study}} \\
\midrule
\multicolumn{2}{c}{} & \multicolumn{7}{c}{\textbf{Class Labels (n = 7)}} \\
\cmidrule(r){3-9}
Data & Source & ER & DF & PS & EC & WQ & FS & SC \\
\midrule
In-domain & \citet{lehmann_manifesto_2024} &  3,149 & 2,133 & 2,588 & 6,902 & 10,501 & 3,412 & 2,131 \\
OOD1 & \citet{osnabrugge2023cross} &  94 & 545 & 1,068 & 720 & 789 & 432 & 325 \\
\bottomrule
\end{tabular}
\caption{\textbf{Descriptive Statistics for Political Topics Study}. ER = External Relations, FD = Freedom and Democracy, PS = Political System, EC = Economy, WQ = Welfare and Quality of Life, FS = Fabric of Society, SC = Social Groups. In-domain train and test datasets were subsequently balanced for the analysis.}
\label{table:political_topics}
\end{table*}

\begin{table}[h!]
\centering
\small
\resizebox{\columnwidth}{!}{%
\begin{tabular}{llcc}
\midrule
\multicolumn{4}{c}{\textbf{Sexism Study}} \\
\toprule
\multicolumn{2}{c}{} & \multicolumn{2}{c}{\textbf{Class Labels (n = 2)}} \\
\cmidrule(r){3-4}
Data & Source & S & nS \\
\midrule
In-domain & \citet{samory2021call} & 1,269 & 11,484 \\
OOD1 & \citet{EXIST2021} & 1,636 & 1,800 \\
OOD2 & \citet{guest2021expert} & 699 & 5,856 \\
OOD3 & \citet{kirk2023semeval} & 4,854 & 15,146 \\
OOD4 & \citet{rottger2020hatecheck} & 373 & 136 \\
\bottomrule
\end{tabular}%
}
\caption{\textbf{Descriptive Statistics for Sexism Study}. S = Sexist, nS = non-Sexist. In-domain train and test datasets were subsequently balanced for the analysis.}
\label{table:sexism_study}
\end{table}

\subsubsection{Sexism Study}

For the in-domain dataset, we use the dataset curated by \citet{samory2021call}, which has multiclass categorization of sexism. However, since some of the categories have few labels, we instead use the binary sexism labels.  

For the OOD evaluation, we use other English-language sexism datasets with binary sexism labels, specifically the EDOS dataset \citet{kirk2023semeval}, the Reddit Misogyny dataset \citet{guest2021expert}, and the EXIST shared task dataset \citet{EXIST2021}, and finally a subset of the Hatecheck test suite that targets women \citet{rottger2020hatecheck}. 

\subsubsection{Political Topics Study}

For the multi-class dataset, we rely on a dataset of annotated sentences from UK party manifestos between 1964 and 2019 gathered from the Party Manifesto Project \citep{lehmann_manifesto_2024}.
Sentences are coded into one of seven political issues, that are \textit{external relations, freedom and democracy, political system, economy, welfare and quality of life, fabric of society} and \textit{social groups}. 
As an OOD dataset, we rely on a dataset by \citet{osnabrugge2023cross} who coded 4,165 New Zealand parliamentary speeches using the same codebook and classes. Hence, the labels should be comparable despite utilising a different text source and political context in a foreign parliamentary system. See Appendix \ref{topics_detailed} for an overview of all the political topics and their descriptions. 

\subsection{Experimental Setup}

\begin{figure}[h]
    \centering
    \includegraphics[width=1\linewidth]{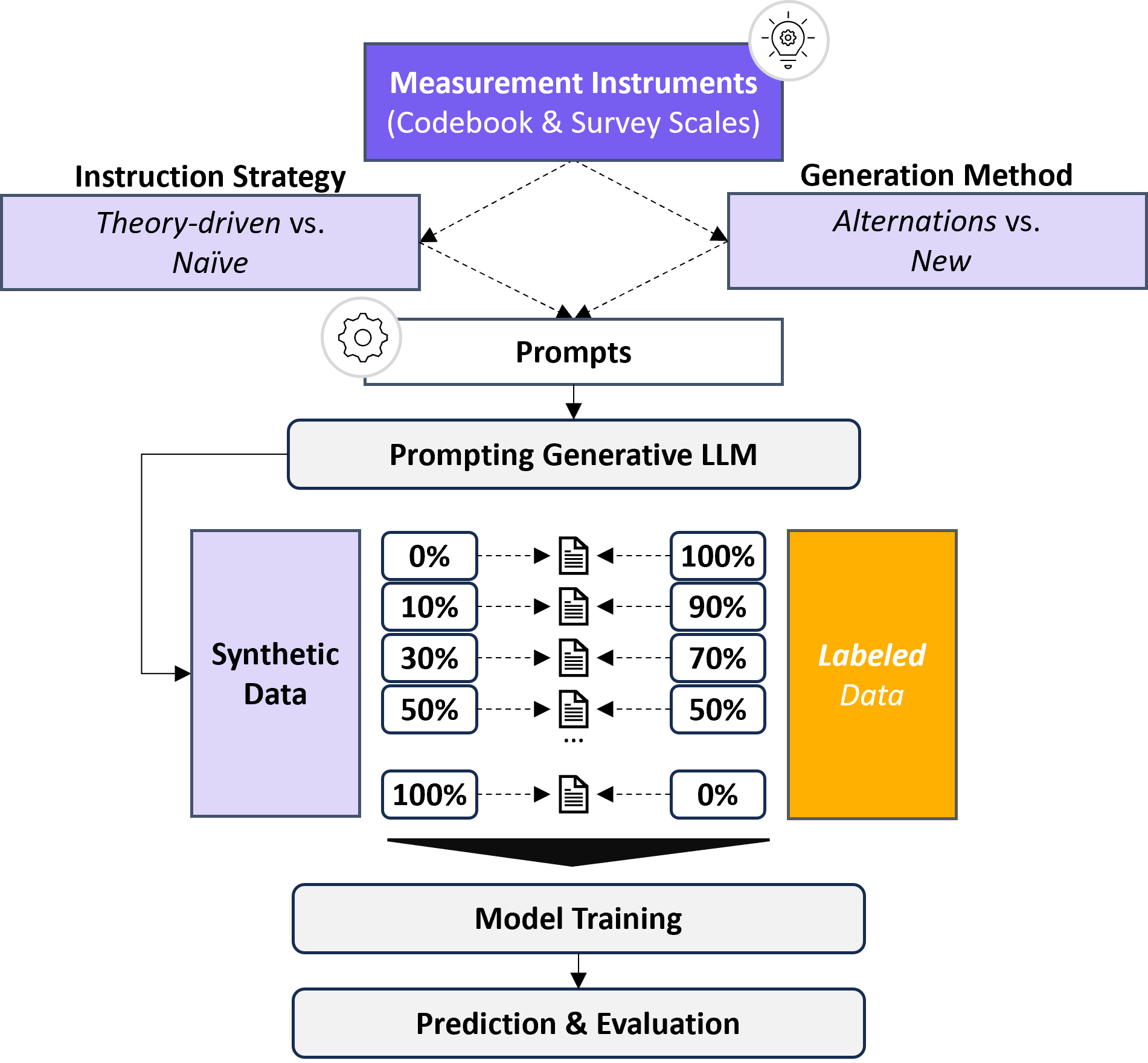}
    \caption{\textbf{Computational workflow} to generate training datasets that contain different shares of synthetic and human-labeled data}
    \label{fig:workflow}
\end{figure}

Figure \ref{fig:workflow} visualizes our workflow. 

\subsubsection{Synthetic Data Generation}

Based on the different prompting strategies identified in Table \ref{tab:overview}, we generate different types of synthetic data. The generation of synthetic data was varied according to two factors: \textbf{instruction strategy} (\textit{theory-driven} or \textit{naive}) and \textbf{generation method} (\textit{alternations of existing data} or \textit{newly generated data}). 
Both combinations of prompting strategies allow us to obtain labeled instances of synthetic data without human labeling. Below, we further outline the basic ideas behind our taxonomy.  See Appendix \ref{sec:appendix_prompts}  for the full overview of all prompts used to generate synthetic data according to the combination of both variants. 

\textbf{Instruction Strategy:} We differentiate between \textit{theory-driven} and \textit{naive} strategies for creating synthetic training data. 
For the \textit{theory-driven} strategy, we prompt the model to generate texts in accordance with the conceptual information based on the social science measurement instruments. 
For the sexism study, we prompt the model to generate only sexist tweets pertaining to the phrasing of survey scales initially designed to measure sexist behaviour. \footnote{For sexism, we generated only sexist synthetic data due to the infinite variety of non-sexist examples that could be created.} This decision was made to ensure the model is trained on explicit examples of sexist behavior, increasing its ability to recognize subtle instances of sexism. 
We draw upon the inventory compiled by \citet{samory2021call}, which aggregates multiple survey items into four overarching dimensions: \textit{Behavioral Expectations}, \textit{Stereotypes and Comparative Opinions}, \textit{Endorsement of Inequality}, and \textit{Denying Inequality and Rejecting Feminism} (for a detailed overview, see Appendix \ref{examples}).
For example, the scale items relating to the dimension \textit{Behavioral Expectations} include several statements describing sexist behaviour relating to behavioural expectations, such as "A woman should be careful not to appear smarter than the man she is dating." 

For the political topics study, we instructed the model to generate text related to specific subtopics of the party manifesto codebook. For example, the subtopic "support of traditional morality" under the topic "fabric of Society" is described with details like "suppression of immorality and unseemly behavior," "maintenance and stability of the traditional family," and "support for the role of religious institutions in state and society." 

For the \textit{naive} strategy, the model is instructed to only generate tweets that are sexist (sexism study), or that are about a specific topic (political topics study). Thus, we did not specify any details on how the construct should express itself in textual data and provide no further information beyond the label's name.

\textbf{Generation Type:} We further differentiate between \textit{alternations of existing data} and \textit{newly generated data}. 

For the \textit{newly generated data}, we only provide the models with a prompt to create new texts based on the category of the initial data, that is, social media posts (sexism study) and party manifesto sentences (political topics study). 
For the \textit{alternations of existing data}, we provide the models with examples from the original corpus that should be alternated. For each prompting call, we sample a random text from the raw text corpus and include it in the prompt. Thus, we ask the model to only make minimal changes to the text while rewriting it to be in line with the intended class of the construct of interest (i.e., being sexist (sexism study) or about a specific topic, such as \textit{economy} or {external relations} (political topics study)).  
We then generated a balanced corpus of synthetic data using two different generative models for each unique combination of prompting strategies. The models we used were: 

\textbf{GPT-3.5-turbo.} For both studies, we rely on a GPT-3.5-turbo model with the following default parameters for our API requests: GPT-3.5-turbo, with a temperature of $0.7$.  

\textbf{Llama-3-70b}. For the political topics study, we rely on the pretrained and instruction-tuned Meta-Llama-3.1-70B-Instruct model. For the sexism study, we rely on an uncensored instruct-tuned Llama-3-70b model trained on the \textit{Dolphin} dataset \citep{hartfordDolphinLlama70b}. We take advantage of the uncensored version of the model to generate the sexist tweets as the instruction Llama-3 models refused to generate sexist tweets altogether.  

For both models, we used default parameters with a temperature of $0.7$. We prompted the models to generate five individual texts along with their corresponding labels in JSON format, then converted the output into a tabular format for further processing.

\subsubsection{Model Training}

For each unique combination of prompting strategies, we then create training datasets with varying proportions of synthetic data, ranging from 0\% (only labeled data), 30\%, 50\%, 70\%, 90\%, to 100\% (only synthetic data). We create a balanced training set with 1,938 cases for the sexism study (969 cases per class) and 3,500 cases for the political topics study (500 cases per class). The in-domain test set, which contains no synthetic data, includes 600 cases (300 per class) for the sexism study and 3,500 cases for the political topics study. From this, we randomly sample 30\% as a validation set, with the remaining 70\% serving as the final held-out test set. To ensure reproducibility, a random seed is set for all nondeterministic operations. 

Next, we introduce the models that we use for classification. All hyperparameters and our computational resources are documented in Appendix \ref{hyperpara} and \ref{comp}, respectively.

\textbf{RoBERTa (X):} Our primary models are pretrained RoBERTa models \citep{liu2019roberta} fine-tuned on the full size of the training dataset. However, the training datasets vary in the share of synthetic data (denoted by the value of X). For each prompting strategy, we train models with 0\%, 30\%, 50\%, 70\%, 90\%, and 100\% shares of synthetic data. For instance, the model RoBERTa (70) is a model trained on a dataset that contains 70\% synthetic data and 30\% labeled data. In contrast, the model RoBERTa (0) is trained on a dataset that contains only labeled data (with no share of synthetic data). 

\textbf{RoBERTa (X - labeled subset):} As a baseline, we train RoBERTa (X) models using only the labeled portion of the training data to assess the impact of adding synthetic data. Since these training datasets are smaller than those used for the RoBERTa (X) models with synthetic data, we include them primarily as a robustness check. Performance differences should not be attributed solely to data quality, as they could also be due to the smaller training size.

\textbf{SVM (X):} As a baseline, we also rely on support vector machine models (SVM) trained on the same varying shares of synthetic data as the RoBERTa (X) models. Though simple, SVM is computationally efficient and serves as a solid benchmark for comparison. 

\textbf{Prompting GPT-4o}: Furthermore, in line with current discussions around generative models being used for classification \citep{gilardi2023chatgpt}, we test the performance of our models against an alternative measurement design: asking a generative model to classify the texts directly. While this form of classification comes with its limitations, in particular reliability and reproducibility \citep{reiss2023testing, breuerAreWeReplicating2024}, we still consider it a relevant benchmark due to its growing applicability and potential for applied measurement designs~\cite{atreja2024prompt}. 
Thus, we use GPT-4o for direct labeling with default parameters (see Appendix \ref{prompts} for the full prompts).

\section{Results}

Subsequently, we report the results of the analysis.
\subsection{Overall Performance}

Table \ref{tab:per} and Table \ref{tab:ood} show the best-performing models that were trained on datasets with different shares of synthetic data for in-domain and OOD, respectively. 

\begin{table}[h!]
\centering
\scriptsize
\begin{tabular}{llll|llll}
\toprule
\multicolumn{4}{c|}{Sexism Study} & \multicolumn{4}{c}{Political Topics Study} \\
\midrule
Model & I & G & F1 & Model & I & G &  F1 \\
\midrule
 RoBERTa (0)& - &-& .89  & RoBERTa (0)& - & -&\textbf{.64} \\
 RoBERTa (30) & T & N & .90  & RoBERTa (30) &  T&A & .63 \\
 RoBERTa (50)&  T&A & .90  & RoBERTa (50)& T&N & .62 \\
 RoBERTa (70)&  T&A & .89  & RoBERTa (70)&  T&A & .60 \\
 RoBERTa (90)&  T&A & .81  & RoBERTa (90)&  T&A & .56 \\
 RoBERTa (100)&  N&A & .64  & RoBERTa (100)&  T&A & .52 \\
   SVM (0) &-&- & .85  & SVM (0) &-& -& .49 \\
   SVM (30) &N&A& .85  & SVM (30) &T&A& .49 \\
   SVM (50) &N&A& .84  & SVM (50) &T&A& .48 \\
   SVM (70) &N&A& .81  & SVM (70) &T&A& .46 \\
   SVM (90) &N&A& .77  & SVM (90) &T&A& .44 \\
   SVM (100)& N&A& .72  & SVM (100) &T&A& .38 \\
 Prompting & -& -&\textbf{.91} & Prompting &- &-& .43 \\
\bottomrule
\end{tabular}
\caption{\textbf{Best In-domain Performance}. Column I refers to \textit{Instruction Strategy}, with T = Theory-driven and N = New; Column G refers to the \textit{Generation Type}, with N = Naive and A = Alternations. The prompting model refers to the baseline GPT-4o. For the sexism model, we report average performance across four out-of-domain (OOD) test sets. The best-performing models are highlighted in bold.}
\label{tab:per}
\vspace{-8pt}
\end{table}

\textbf{In-Domain:} Overall, our findings indicate that the inclusion of synthetic data did \textbf{not result in a substantive improvement in the classifier's \textit{overall} performance} on the in-domain test set. 

In the political topics study, the highest F1 score was achieved by the RoBERTa (0) model trained solely on labeled data. As the share of synthetic training data increases, the respective RoBERTa (X) models performed slightly worse. For instance, the RoBERTa (70) model had a macro-F1 score that was 0.04 lower than the RoBERTa (0) model.  

Conversely, in the sexism study, models trained with 30\% and 50\% synthetic data performed marginally better, with a macro-F1 score of .90 compared to .89 for the RoBERTa (0) model, but dropping significantly in performance for larger shares of synthetic data. This suggests that the addition of synthetic data might have contributed minimally to the diversity of examples and patterns in the training data, leading to a slight improvement in the model’s \textit{overall performance} on the test set. 

As expected, the RoBERTa (X) models generally outperformed the respective baseline SVM (X) models, with an overall macro-F1 increase of $+.05$ for the sexism study and $+.15$ for the topics study.

Finally, two key observations emerge when comparing the in-domain performance to the baseline prompting approach. The sexism study achieved the highest performance with a simple prompting strategy, yielding an F1 score of $.91$ versus $.89$ for the RoBERTa (0) model that was trained on only human-labeled data. In the topics study, however, the RoBERTa (0) model clearly surpassed the prompt-based design, with an F1 score of $.64$ compared to $.43$ for the prompt-based approach.

\textbf{OOD:} Overall, our findings (Table~\ref{tab:ood} and Figure~\ref{fig:ood} in the Appendix) indicate that the inclusion of synthetic data did only marginally improve \textit{overall} OOD performance. 

\begin{table}[h!]
\centering
\scriptsize
\begin{tabular}{llll|llll}
\toprule
\multicolumn{4}{c|}{Sexism Study} & \multicolumn{4}{c}{Political Topics Study} \\
\midrule
Model & I & G & F1 & Model & I & G &  F1 \\
\midrule
RoBERTa (0) & - & - & .58  & RoBERTa (0) & - & - & .56 \\
RoBERTa (30) & T & N & .58  & RoBERTa (30) & T & A & \textbf{.57} \\
RoBERTa (50) & T & A & .56  & RoBERTa (50) & T & N & \textbf{.57} \\
RoBERTa (70) & T & A & .54  & RoBERTa (70) & T & A & .56 \\
RoBERTa (90) & N & A & .49  & RoBERTa (90) & T & A & .54 \\
RoBERTa (100) & N & A & .44  & RoBERTa (100) & T & A & .48 \\
SVM (0) & - & - & .46  & SVM (0) & - & - & .28 \\
SVM (30) & N & A & .46  & SVM (30) & T & A & .28 \\
SVM (50) & N & A & .45  & SVM (50) & T & A & .29 \\
SVM (70) & N & A & .44  & SVM (70) & T & A & .30 \\
SVM (90) & N & A & .42  & SVM (90) & T & A & .28 \\
SVM (100) & N & A & .40  & SVM (100) & T & A & .28 \\
     Prompting      &-& - & \textbf{.60} & Prompting   &-&  -   &  .34 \\
\bottomrule
\end{tabular}
\caption{\textbf{Best OOD Performance}. Column I refers to \textit{Instruction Strategy}, with T = Theory-driven and N = New; Column G refers to the \textit{Generation Type}, with N = Naive and A = Alternations. The prompting model refers to the baseline GPT-4o. For the sexism model, we report average performance across four out-of-domain (OOD) test sets. The best-performing models are highlighted in bold.}
\label{tab:ood}
\vspace{-8pt}
\end{table}

For the political topics study, RoBERTa (X) models trained with 30\% and 50\% synthetic data exhibited marginally improved performance, maintaining stable results up to 90\% synthetic data. Conversely, the sexism study demonstrated relatively lower but still stable performance across models utilizing synthetic data; however, these models generally underperformed compared to the RoBERTa (0) model that was exclusively trained on labeled data.

When assessing the baseline prompting method, the sexism study indicates slightly higher performance (F1 = .60) than the RoBERTa (X) models. In the political topics study, the prompting approach performed significantly worse overall, reinforcing that semi-supervised classification outperformed prompt-based classification for political topics classification.

\subsection{Effect of Prompting Strategy}

While our analysis suggests that adding synthetic training data did not lead to a notable improvement in \textit{overall} performance, we further investigate how integrating fractions of varying synthetic data into the training data can reduce the amount of human-labeled data required by serving as a partial substitute.

\begin{figure*}[h!]
    \centering
    \includegraphics[width=0.9\linewidth]{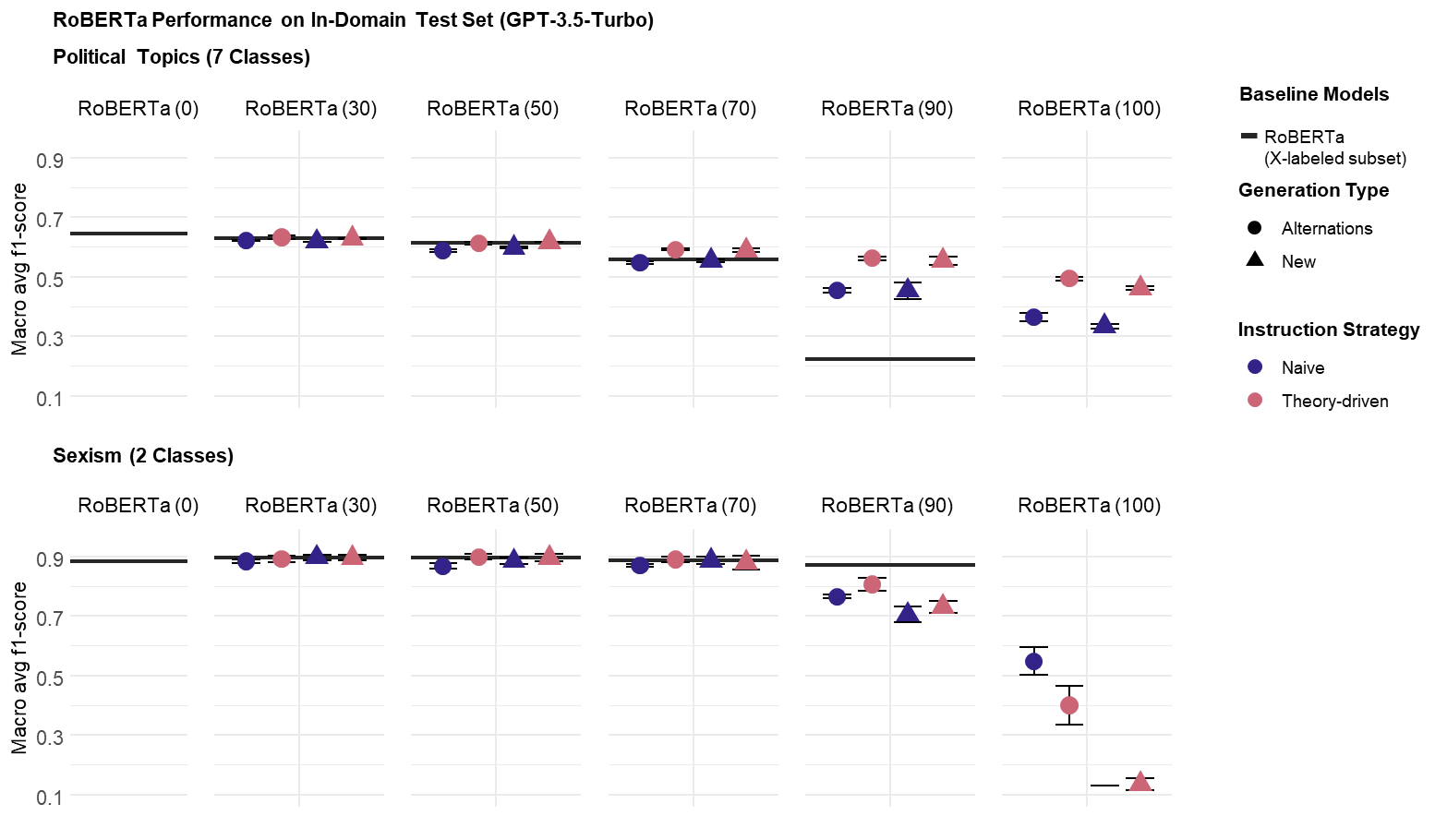}
    \caption{\textbf{Performance on In-Domain Test Set}. The y-axis depicts the macro-F1 score across three random seeds. Error bars represent 95\% confidence intervals across the three training runs. Training sizes were balanced across classes, with $N_{Training}$ = 1,938 for the sexism study (969 cases per class) and $N_{Training}$ = 3,500 for the topics study (500 cases per class). The dark grey lines correspond to the results of a model trained on only the labeled subset of the data; synthetic data was generated using GPT-3.5-turbo. }
    \label{fig:res_1}
\end{figure*}

\textbf{In-Domain:} Figure \ref{fig:res_1} displays the in-domain model performances across the varying prompting strategies for both studies.
%Plots
%Plot Description
The figure visualises the mean performance (average macro-F1) across three random seeds for the RoBERTa (X) models. Each column represents the results for RoBERTa models trained on the same subsets of synthetic and human-labeled data for different types of synthetic data. The color and shape of the points reflect the prompting strategies used to generate the synthetic data. For the \textit{instruction strategy}, blue points represent "theory-driven" and red points represent "naive" strategies. For the \textit{generation type}, dots indicate "alternations" of the original data, while triangles represent "new" data generated without any text examples. The dark grey line in each column corresponds to the performance of the RoBERTa (X-labeled subset) model, the model trained on only the labeled subset of the training data. For instance, the model RoBERTa (70-labeled subset) corresponds to the performance of a RoBERTa model that is only trained on the remaining 30\% of the human-labeled data (with the 70\% synthetic data being removed from the training data).  

Overall, our results are twofold.
%Topic Study
%% Overall
In the political topics study, \textbf{we observe a consistent, slight decline in performance as the proportion of synthetic data increases across different prompting strategies}. 
Nonetheless, our results demonstrate that a model trained with just 30 per cent labeled data and 70 per cent synthetic data achieved an F1-score of .60, only $.04$ lower than a model trained on only human-labeled data. A model trained exclusively on synthetic data still attained a respectable macro-F1 score of .52.  Furthermore, most RoBERTa (X) models achieve significantly higher performance than their RoBERTa (X-labeled subset) baseline models, indicating that the addition of synthetic data helped the model in the training process beyond the labeled data available in the training dataset. 
In contrast, for the sexism study, including synthetic data did not improve model performance. Across all data subsets, the RoBERTa (X-labeled subset) models trained on only the share of labeled data performed equally well or even outperformed those trained on the combined labeled and synthetic subsets.
Figure \ref{fig:ood} in Appendix \ref{perfo_ood} illustrates that this pattern is consistent for OOD classification.

To further investigate the impact of a theory-driven data generation strategy, Table \ref{table:diff_gen} presents the mean differences between models trained on theory-driven data and those trained on synthetic data generated using a naive \textit{instruction strategy}.
\begin{table}[h!]
\centering
\small 
\begin{tabular}{lcc}
\midrule
\multicolumn{3}{c}{\textbf{Political Topics Study}} \\
\toprule
Model &  Diff In-Domain & Diff OOD  \\
\midrule
RoBERTa (30) &    \textbf{+.03}&\textbf{+.01} \\
RoBERTa (50)&  \textbf{+.03} &  \textbf{+.02}\\
RoBERTa (70) &  \textbf{+.06}&  \textbf{+.04} \\
RoBERTa (90)  &  \textbf{+.11}&  \textbf{+.10}\\
RoBERTa (100)  &  \textbf{+.12}&  \textbf{+.14}\\
\midrule
\multicolumn{3}{c}{\textbf{Sexism Study}} \\
\toprule
Model &  Diff In-Domain & Diff OOD  \\
\midrule
RoBERTa (30) &  \textbf{+.00}&   \textbf{+.00} \\
RoBERTa (50) &  \textbf{+.02}&   \textbf{+.02} \\
RoBERTa (70) &  \textbf{+.02}& \textbf{+.02} \\
RoBERTa (90)  &  \textbf{+.00}&  \textbf{+.00}\\
RoBERTa (100) &  \textbf{+.00}&  \textbf{-.10} \\
\toprule
\end{tabular}
\caption{\textbf{Impact of Theory-Driven vs. Naive Instruction Strategy}. 'Diff' shows the performance gap between models trained on theory-driven and naive synthetic data, with positive values indicating better performance of theory-driven synthetic training data.}
\label{table:diff_gen}
\vspace{-8pt}
\end{table}
In the political topics study, \textbf{models trained on theory-driven data consistently outperformed the naive ones for both in-domain and OOD classification}. As the proportion of synthetic data increased, the performance gap between the two prompting strategies widened, reaching a difference for the  RoBERTa (100) models of up to 14\% for the in-domain and 12\% for the OOD test sets.
In the sexism study, synthetic data generally did not benefit the models. This is evident in the lack of significant differences across prompting strategies, where models trained on theory-driven data showed only slightly better performance on the in-domain dataset but worse performance on the OOD dataset (see Table \ref{table:diff_gen}).

In Appendix \ref{robstuness_checks}, we discuss the results from using Llama to generate the synthetic training data and find an overall lower performance but similar patterns, indicating that our findings are robust against the generation model.

\section{Discussion and Conclusion}

In this study, we systematically assessed the effect of including theory-driven synthetic data in the training process to improve the measurement of social constructs. Using two studies, we systematically varied factors relevant to generating theory-driven synthetic data. The results were ambivalent.
In the sexism study, there was no benefit in replacing part of the labeled data with synthetic data. In contrast, for classifying political topics, we observed a clear benefit of replacing parts of the labeled data with synthetic data.

This could be due to several reasons. For one, our topics study included a multi-class setting with seven output classes, thus requiring more varied and diverse examples to capture the nuances of each class. The introduction of theory-driven synthetic data may have helped supplement the variability that was missing in the labeled dataset, allowing the model to generalize better across the different topics. This result mirrors the conclusions of \citet{moller2023parrot} who conclude that synthetic data can improve multi-label classification models if class imbalance is apparent in the original dataset. On the other hand, for sexism, we used theory to generate training data for the sexist class only, since it is not possible to generate "non-sexist" instances in this paradigm. Other types of synthetic augmentation, e.g., counterfactual augmentations \citet{sen2023people}, could be more suitable for sexism detection.

These results have several implications. First, theory-driven sources from social science measurement instruments, in particular annotation codebooks, can help produce better theory-driven synthetic data. Still, this approach did not work well for survey scales. Thus, more research is needed to determine which measurement instruments or frameworks could furthermore be utilized beyond the ones used in our studies. Second, synthetic data could be particularly useful in scenarios with low-resource classes or when many classes are involved, allowing for the replacement of labeled data without severe performance loss. These findings were robust across different models and generation strategies.

\section{Limitations}

For our study design, several limitations must be acknowledged. First, we focus on only two constructs where the theoretical sources (i.e., the codebook and survey scales) diverge. This narrow scope may contribute to inconsistent results across our two studies. Specifically, one might question whether the concepts captured in the survey scales (e.g., behavioral expectations) are adequately represented in the sexist dataset. Alternatively, other features may have played a more prominent role in differentiating the constructs, potentially overshadowing the intended focus on the survey scales.

Second, we rely on a single OOD dataset for evaluation in the political topics study. This limitation is due to the lack of available OOD datasets for multiclass classification in this specific domain. However, using only one dataset restricts the generalizability of our findings and limits our ability to draw broader conclusions about the model's effectiveness for OOD  classification. Future research could benefit from incorporating more OOD datasets for a more thorough evaluation of political topic classification.

Finally, we only examine two model architectures to generate the synthetic data, i.e., GPT-3.5-turbo and Llama 3-70b. Different architectures might behave differently, especially when applied to different types of tasks. By exploring a broader selection of model architectures in future work, we could better assess the consistency of the observed trends and ensure that the results hold across various models.

\section{Ethics Statement}
\label{ehitcs}

This study focuses on generating synthetic training data strictly for experimental purposes. We want to be clear that we reject any form of using this data to degrade or harm individuals or groups based on their nationality, ethnicity, religion, or sexual orientation. At the same time, we understand that concerns about the potential misuse of such human-like augmented data for harmful purposes are valid and should not be overlooked.

Regarding the risks associated with our work, synthetic data can pose potential harm to third parties, e.g., in terms of hate speech. While we recognize the inherent risks in synthetic data content, our study does not add to the development or access to existing methods. Instead, it aims to enhance the measurement and prevent such content in real-world scenarios, e.g., by improving content filters.

We used the AI assistant ChatGPT 4o from OPENAI exclusively for code debugging and text rephrasing. All AI-generated outputs were carefully reviewed, and we affirm that all significant contributions in this paper are the original work of the authors.

\section*{Open Science Statements}
anonymized

\bibliography{custom}
\onecolumn

\begin{appendices}

\section{Experimental conditions of related work}
\label{exp}
\begin{table}[h]
\centering
\small
\resizebox{\textwidth}{!}{%
\begin{tabular}{lccccccclc}  
\hline
Studies& Constructs& Tasks& Theory driven prompt& Examples& Alternations& Prompt strategy&  Human + Synth&OOD Test& LLM Direct\\
\hline
Achmann-Denkler et al. 2024& No& No& No& Yes& Yes& No&  Yes&No& Yes\\
Ghanadian et al. 2024& No& Yes& Yes& Yes& -& Yes&  Yes&No& No\\
Hui et al. 2024& No& No& No& Yes& -& Yes&  No&No& No\\
Li et al. 2023& Yes& Yes& No& Yes& Yes& Yes&  No&No& Yes\\
Mahamoudi et al. 2024& Yes& No& No& Yes& Yes& No&  No&No& Yes\\
Moller et al. 2023& Yes& Yes& No& Yes& Yes& No&  No&No& Yes\\
Veselovsky et al. 2023& No& No& No& Yes& Yes& Yes&  No&No& Yes\\
Wagner et al. 2024& No& No& No& No& No& No&  Yes&No&No\\
\hline
This study& Yes& Yes& Yes& Yes& Yes& Yes&  Yes&Yes&Yes\\
\end{tabular}%
}
\caption{Comparison of Studies on Prompting and Classification Tasks. 
\textit{Constructs} refers to whether different constructs were evaluated. \textit{Task} indicates whether the classification task varied. \textit{Theory} shows if the study used theory-driven prompts. \textit{Examples} refers to whether examples were included in prompts. Alternations show if the prompt used included instructions to rewrite a given example. \textit{Prompt strategy} indicates if prompt strategies were varied. \textit{Human + Synth.} refers to the combination of human and synthetic data. \textit{OOD Test} signifies out-of-domain testing, and \textit{LLM Direct} compares classification using LLM without prompting.  "-" indicates that the exact prompt
could not be determined from the information provided by the authors.}
\label{tab:comparison_studies}
\end{table}

\section{Political Topics}
\label{topics_detailed}
\onecolumn

\begin{landscape} % Begin landscape orientation

\begin{adjustwidth}{-2.5 cm}{-2.5 cm}\centering

\scriptsize
\begin{longtable}{p{3cm} p{1.5cm} p{3cm} p{8cm} p{1.5cm}}\toprule
domain\_name &policy\_code &policy\_name &description\_md &domain\_code \\\midrule
\endfirsthead
\caption[Topic Categories]{(continued)}\\
\toprule
domain\_name &policy\_code &policy\_name &description\_md &domain\_code \\\midrule
\endhead
\midrule \multicolumn{5}{r}{\textit{Continued on next page}} \\
\endfoot
\bottomrule
\endlastfoot
Uncoded &0 & per cent of uncoded quasi-sentences (000) &Share of uncoded sentences (quasi-sentences coded with 000) For Manifesto Project Dataset: - Missing information for most documents of Sweden 1949-1982 and Norway 1945-1989 - For all documents that have been coded with version 5 of the Coding Instructions this category is calculated as the sum of 000 codes, per202\_2, per605\_2, and per703\_2 to ensure comparability with data produced with version 4 of the coding instructions. For Manifesto Project South America Dataset: - share of quasi-sentences coded with 000 (without the shares of 202\_2, 605\_2 and 703\_2 codes) &0 \\
External Relations &101 &foreign special + &Favourable mentions of particular countries with which the manifesto country has a special relationship; the need for co-operation with and/or aid to such countries. &1 \\
External Relations &102 &foreign special - &Negative mentions of particular countries with which the manifesto country has a special relationship. &1 \\
External Relations &103 &anti-imperalism &Negative references to imperial behaviour and/or negative references to one state exerting strong influence (political, military or commercial) over other states. May also include: - Negative references to controlling other countries as if they were part of an empire; - Favourable references to greater self-government and independence for colonies; - Favourable mentions of de-colonisation.  &1 \\
External Relations &104 &military + &The importance of external security and defence. May include statements concerning: - The need to maintain or increase military expenditure; - The need to secure adequate manpower in the military; - The need to modernise armed forces and improve military strength; - The need for rearmament and self-defence; - The need to keep military treaty obligations. &1 \\
External Relations &105 &military - &Negative references to the military or use of military power to solve conflicts. References to the ‘evils of war’. May include references to: - Decreasing military expenditures; - Disarmament; - Reduced or abolished conscription. &1 \\
External Relations &106 &peace &Any declaration of belief in peace and peaceful means of solving crises – absent reference to the military. May include: - Peace as a general goal; - Desirability of countries joining in negotiations with hostile countries; - Ending wars in order to establish peace. &1 \\
External Relations &107 &internationalism + &Need for international co-operation, including co-operation with specific countries other than those coded in 101. May also include references to the: - Need for aid to developing countries; - Need for world planning of resources; - Support for global governance; - Need for international courts; - Support for UN or other international organisations. &1 \\
External Relations &108 &europe + &Favourable mentions of European Community/Union in general. May include the: - Desirability of the manifesto country joining (or remaining a member); - Desirability of expanding the European Community/Union; - Desirability of increasing the ECs/EUs competences; - Desirability of expanding the competences of the European Parliament. &1 \\
External Relations &109 &internationalism - &Negative references to international co-operation. Favourable mentions of national independence and sovereignty with regard to the manifesto country’s foreign policy, isolation and/or unilateralism as opposed to internationalism. &1 \\
External Relations &110 &europe - &Negative references to the European Community/Union. May include: - Opposition to specific European policies which are preferred by European authorities; - Opposition to the net-contribution of the manifesto country to the EU budget. &1 \\
Freedom and Democracy &201 &freedom and human rights &Favourable mentions of importance of personal freedom and civil rights in the manifesto and other countries. May include mentions of: - The right to the freedom of speech, press, assembly etc.; - Freedom from state coercion in the political and economic spheres; - Freedom from bureaucratic control; - The idea of individualism. &2 \\
Freedom and Democracy &202 &democracy &Favourable mentions of democracy as the “only game in town”. General support for the manifesto country’s democracy. May also include: - Democracy as method or goal in national, international or other organisations (e.g. labour unions, political parties etc.); - The need for the involvement of all citizens in political decision-making; - Support for either direct or representative democracy; - Support for parts of democratic regimes (rule of law, division of powers, independence of courts etc.) &2 \\
Freedom and Democracy &203 &constitution + &Support for maintaining the status quo of the constitution. Support for specific aspects of the manifesto country’s constitution. The use of constitutionalism as an argument for any policy. &2 \\
Freedom and Democracy &204 &constitution - &Opposition to the entirety or specific aspects of the manifesto country’s constitution. Calls for constitutional amendments or changes. May include calls to abolish or rewrite the current constitution. &2 \\
Political System &301 &decentralisation &Support for federalism or decentralisation of political and/or economic power. May include: - Favourable mentions of the territorial subsidiary principle; - More autonomy for any sub-national level in policy making and/or economics, including municipalities; - Support for the continuation and importance of local and regional customs and symbols and/or deference to local expertise; - Favourable mentions of special consideration for sub-national areas. &3 \\
Political System &302 & centralisation &General opposition to political decision-making at lower political levels. Support for unitary government and for more centralisation in political and administrative procedures. &3 \\
Political System &303 &gov-admin efficiency &Need for efficiency and economy in government and administration and/or the general appeal to make the process of government and administration cheaper and more efficient. May include: - Restructuring the civil service; - Cutting down on the civil service; - Improving bureaucratic procedures.  &3 \\
Political System &304 &political corruption &Need to eliminate political corruption and associated abuses of political and/or bureaucratic power. Need to abolish clientelist structures and practices. &3 \\
Political System &305 &political authority &References to the manifesto party’s competence to govern and/or other party’s lack of such competence. Also includes favourable mentions of the desirability of a strong and/or stable government in general. &3 \\
Economy &401 &free market economy &Favourable mentions of the free market and free market capitalism as an economic model. May include favourable references to: - Laissez-faire economy; - Superiority of individual enterprise over state and control systems; - Private property rights; - Personal enterprise and initiative; - Need for unhampered individual enterprises. &4 \\
Economy &402 &incentives &Favourable mentions of supply side oriented economic policies (assistance to businesses rather than consumers). May include: - Financial and other incentives such as subsidies, tax breaks etc.; - Wage and tax policies to induce enterprise; - Encouragement to start enterprises. &4 \\
Economy &403 &market regulation &Support for policies designed to create a fair and open economic market. May include: - Calls for increased consumer protection; - Increasing economic competition by preventing monopolies and other actions disrupting the functioning of the market; - Defence of small businesses against disruptive powers of big businesses; - Social market economy. &4 \\
Economy &404 &economic planning &Favourable mentions of long-standing economic planning by the government. May be: - Policy plans, strategies, policy patterns etc.; - Of a consultative or indicative nature. &4 \\
Economy &405 &corporatism/mixed economy &Favourable mentions of cooperation of government, employers, and trade unions simultaneously. The collaboration of employers and employee organisations in overall economic planning supervised by the state.  &4 \\
Economy &406 &protectionism + &Favourable mentions of extending or maintaining the protection of internal markets (by the manifesto or other countries). Measures may include: - Tariffs; - Quota restrictions; - Export subsidies. &4 \\
Economy &407 &protectionism - &Support for the concept of free trade and open markets. Call for abolishing all means of market protection (in the manifesto or any other country). &4 \\
Economy &408 &economic goals &Broad and general economic goals that are not mentioned in relation to any other category. General economic statements that fail to include any specific goal.  &4 \\
Economy &409 &keynesian demand management &Favourable mentions of demand side oriented economic policies (assistance to consumers rather than businesses). Particularly includes increase private demand through - Increasing public demand; - Increasing social expenditures. May also include: - Stabilisation in the face of depression; - Government stimulus plans in the face of economic crises. &4 \\
Economy &410 &economic growth + &The paradigm of economic growth. Includes: - General need to encourage or facilitate greater production; - Need for the government to take measures to aid economic growth. &4 \\
Economy &411 &technology and infrastructure &Importance of modernisation of industry and updated methods of transport and communication. May include: - Importance of science and technological developments in industry; - Need for training and research within the economy (This does not imply education in general (see category 506); - Calls for public spending on infrastructure such as roads and bridges; - Support for public spending on technological infrastructure (e.g.: broadband internet, etc.). &4 \\
Economy &412 &controlled economy &Support for direct government control of economy. May include, for instance: - Control over prices; - Introduction of minimum wages. &4 \\
Economy &413 &nationalisation &Favourable mentions of government ownership of industries, either partial or complete; calls for keeping nationalised industries in state hand or nationalising currently private industries. May also include favourable mentions of government ownership of land. &4 \\
Economy &414 &economic orthodoxy &Need for economically healthy government policy making. May include calls for: - Reduction of budget deficits; - Retrenchment in crisis; - Thrift and savings in the face of economic hardship; - Support for traditional economic institutions such as stock market and banking system; - Support for strong currency. &4 \\
Economy &416 &anti-growth economy + &Favourable mentions of anti-growth politics. Rejection of the idea that all growth is good growth. Opposition to growth that causes environmental or societal harm. Call for sustainable economic development.  &4 \\
Welfare and Quality of Life &501 &environmentalism + &General policies in favour of protecting the environment, fighting climate change, and other “green” policies. For instance: - General preservation of natural resources; - Preservation of countryside, forests, etc.; - Protection of national parks; - Animal rights. May include a great variance of policies that have the unified goal of environmental protection. &5 \\
Welfare and Quality of Life &502 &culture + &Need for state funding of cultural and leisure facilities including arts and sport. May include: - The need to fund museums, art galleries, libraries etc.; - The need to encourage cultural mass media and worthwhile leisure activities, such as public sport clubs. &5 \\
Welfare and Quality of Life &503 &equality + &Concept of social justice and the need for fair treatment of all people. This may include: - Special protection for underprivileged social groups; - Removal of class barriers; - Need for fair distribution of resources; - The end of discrimination (e.g. racial or sexual discrimination). &5 \\
Welfare and Quality of Life &504 &welfare + &Favourable mentions of need to introduce, maintain or expand any public social service or social security scheme. This includes, for example, government funding of: - Health care - Child care - Elder care and pensions - Social housing  &5 \\
Welfare and Quality of Life &505 &welfare - &Limiting state expenditures on social services or social security. Favourable mentions of the social subsidiary principle (i.e. private care before state care); &5 \\
Welfare and Quality of Life &506 &education + &Need to expand and/or improve educational provision at all levels.  &5 \\
Welfare and Quality of Life &507 &education - &Limiting state expenditure on education. May include: - The introduction or expansion of study fees at all educational levels - Increasing the number of private schools. &5 \\
Fabric of Society &601 &national way of life + &Favourable mentions of the manifesto country’s nation, history, and general appeals. May include: - Support for established national ideas; - General appeals to pride of citizenship; - Appeals to patriotism; - Appeals to nationalism; - Suspension of some freedoms in order to protect the state against subversion.  &6 \\
Fabric of Society &602 &national way of life - &Unfavourable mentions of the manifesto country’s nation and history. May include: - Opposition to patriotism; - Opposition to nationalism; - Opposition to the existing national state, national pride, and national ideas. &6 \\
Fabric of Society &603 &traditional morality + &Favourable mentions of traditional and/or religious moral values. May include: - Prohibition, censorship and suppression of immorality and unseemly behaviour; - Maintenance and stability of the traditional family as a value; - Support for the role of religious institutions in state and society. &6 \\
Fabric of Society &604 &traditional morality - &Opposition to traditional and/or religious moral values. May include: - Support for divorce, abortion etc.; - General support for modern family composition; - Calls for the separation of church and state. &6 \\
Fabric of Society &605 &law and order + &Favourable mentions of strict law enforcement, and tougher actions against domestic crime. Only refers to the enforcement of the status quo of the manifesto country’s law code. May include: - Increasing support and resources for the police; - Tougher attitudes in courts; - Importance of internal security.  &6 \\
Fabric of Society &606 &civic mindedness + &Appeals for national solidarity and the need for society to see itself as united. Calls for solidarity with and help for fellow people, familiar and unfamiliar. May include: - Favourable mention of the civil society; - Decrying anti-social attitudes in times of crisis; - Appeal for public spiritedness; - Support for the public interest.  &6 \\
Fabric of Society &607 &multiculturalism + &Favourable mentions of cultural diversity and cultural plurality within domestic societies. May include the preservation of autonomy of religious, linguistic heritages within the country including special educational provisions. &6 \\
Fabric of Society &608 &multiculturalism - &The enforcement or encouragement of cultural integration. Appeals for cultural homogeneity in society.  &6 \\
Social Groups &701 &labour groups + &Favourable references to all labour groups, the working class, and unemployed workers in general. Support for trade unions and calls for the good treatment of all employees, including: - More jobs; - Good working conditions; - Fair wages; - Pension provisions etc. &7 \\
Social Groups &702 &labour groups - &Negative references to labour groups and trade unions. May focus specifically on the danger of unions ‘abusing power’. &7 \\
Social Groups &703 &agriculture + &Specific policies in favour of agriculture and farmers. Includes all types of agriculture and farming practises. Only statements that have agriculture as the key goal should be included in this category. &7 \\
Social Groups &704 &middle class and prof. groups &General favourable references to the middle class. Specifically, statements may include references to: - Professional groups, (e.g.: doctors or lawyers); - White collar groups, (e.g.: bankers or office employees), - Service sector groups (e.g.: IT industry employees); - Old and/or new middle class.  &7 \\
Social Groups &705 &minority groups &Very general favourable references to underprivileged minorities who are defined neither in economic nor in demographic terms (e.g. the handicapped, homosexuals, immigrants, indigenous). Only includes favourable statements that cannot be classified in other categories (e.g. 503, 504, 604, 607 etc.) &7 \\
Social Groups &706 &non-economic demographic groups &General favourable mentions of demographically defined special interest groups of all kinds. They may include: - Women; - University students; - Old, young, or middle aged people. Might include references to assistance to these groups, but only if these do not fall under other categories (e.g. 503 or 504). &7 \\

\end{longtable}
\label{topics}
\end{adjustwidth}

\end{landscape} % End landscape orientation

\section{Prompts Synthetic Data Generation}
\label{sec:appendix_prompts}

\begin{table*}[!htbp] 
    \centering
    \small
    \begin{tabularx}{\textwidth}{|p{2cm}|p{2cm}|X|X|}
         \hline
         \textbf{Instruction Strategy} & \textbf{Generation Type} & \textbf{Role Description} & \textbf{User Message} \\
         \hline
         Theory-driven & New & \textbf{Role:} You are a social media user. Generate 5 exemplary tweets for a person that would score high on the following sexism survey item, therefore supporting the statement. The tweets should reflect the underlying concepts of the survey scale, demonstrating attitudes and beliefs that align with high sexism scores. & \textbf{User:} The survey item is: \textit{[Random Survey 
 Item]}. Only display the list of the 5 new tweets and put each tweet in quotation marks. \\
         \hline
         Theory-driven & Alternations & \textbf{Role:} You are a social media user. Rewrite the following exemplary tweets for a person that would score high on the following sexism survey item, therefore supporting the statement. The tweets should reflect the underlying concepts of the survey scale, demonstrating attitudes and beliefs that align with high sexism scores. Keep the same style and slang of the examples and only do necessary changes. & \textbf{User:} The survey item is: \textit{[Random Survey Item]}. Examples to rewrite are: \textit{[List of 5 randomly selected tweets]}. Only display the list of the 5 new tweets and put each tweet in quotation marks. \\
         \hline
         Naive & New & \textbf{Role:} You are a social media user. Generate 5 exemplary sexist tweets. & \textbf{User:} Only display the list of the 5 new tweets and put each tweet in quotation marks. \\
         \hline
         Naive & Alternations & \textbf{Role:} You are a social media user. Rewrite the following exemplary tweets to make them sexist. The tweets should reflect a sexist behavior. Keep the same style and slang of the examples and only do necessary changes. & \textbf{User:} Examples to rewrite are: \textit{[List of 5 randomly selected tweets]}. Only display the list of the 5 new tweets and put each tweet in quotation marks. \\
         \hline
    \end{tabularx}
        \caption{Prompt Designs for Sexism Study}

    \label{tab:my_label}
\end{table*}

\begin{table*}[!htbp] 
    \centering
    \small
    \begin{tabularx}{\textwidth}{|p{2cm}|p{2cm}|X|X|}
         \hline
         \textbf{Instruction Strategy} & \textbf{Generation Type} & \textbf{Role Description} & \textbf{User Message} \\
         \hline
         Theory-driven & New & \textbf{Role:} You are a party manager responsible for drafting your party manifesto. Generate 5 exemplary sentences that could appear in a manifesto about the topic \textit{[topic]}. & \textbf{User:} Specifically create sentences about the subtopic -\textit{[subtopic]}-, which is defined as: \textit{[topic description]}. Only display the list of the 5 new sentences and put each sentence in quotation marks. \\
         \hline
         Theory-driven & Alternations & \textbf{Role:} You are a party manager responsible for drafting your party manifesto. Rewrite the following sentences to make them about the topic \textit{[topic]}. Keep the same style and tone of the example and only make necessary changes. & \textbf{User:} Specifically create sentences about the subtopic -\textit{[subtopic]}-, which is defined as: \textit{[topic description]}. Example to rewrite is: \textit{[Example sentence]}. Only display the list of the 5 new sentences and put each sentence in quotation marks. \\
         \hline
         Naive & New & \textbf{Role:} You are a party manager responsible for drafting your party manifesto. Generate 5 exemplary sentences that could appear in a manifesto about the topic \textit{[topic]}. & \textbf{User:} Only display the list of the 5 new sentences and put each sentence in quotation marks. \\
         \hline
         Naive & Alternations & \textbf{Role:} You are a party manager responsible for drafting your party manifesto. Rewrite the following sentences to make them about the topic \textit{[topic]}. Keep the same style and tone of the example and only make necessary changes. & \textbf{User:} Example to rewrite is: \textit{[Example sentence]}. Only display the list of the 5 new sentences and put each sentence in quotation marks. \\
         \hline
    \end{tabularx}
    \caption{Prompt Design for Political Topics Study}
    \label{tab:prompt_designs}
\end{table*}

\section{Example Survey Items}
\label{examples}
\begin{table*}[!htbp]
\centering
\small
\renewcommand{\arraystretch}{2} % Adjust the value to increase/decrease space between rows
\begin{tabular}{p{2.5cm}p{7cm}p{4cm}} \hline
\textbf{Category} & \textbf{Definition} & \textbf{Scale Item} \\\hline
\textbf{Behavioral \newline Expectations} & Items formulating a \textit{prescriptive} set of behaviors or qualities, that women (and men) are supposed to exhibit in order to conform to traditional gender roles & \textit{"A woman should willingly take her husband's name at marriage"} \\
\textbf{Stereotypes and Comparative Opinions} & Items formulating a \textit{descriptive} set of properties that supposedly differentiates men and women. Those supposed differences are expressed through explicit comparisons and stereotypes. & \textit{"Men make better engineers than women"} \\
\textbf{Endorsement of Inequality} & Items acknowledging inequalities between men and women but justifying or endorsing these inequalities. & \textit{"I think boys should be brought up differently than girls"} \\
\textbf{Denying \newline Inequality and Rejecting Feminism} & Items stating that there are no inequalities between men and women (any more) and/or that they are opposing feminism & \textit{"Discrimination against women is no longer a problem in the United States"} \\
\hline
\end{tabular}
\caption{\textbf{Exemplary Survey Items for Sexism}}
\label{prompts}
\end{table*}

\newpage

\section{Prompts Classification}
\label{prompts}

Costs for the prompt-based classification were around 15\$. 

\begin{table*}[!htbp] 
    \centering
        \small

    \begin{tabularx}{\textwidth}{|p{2cm}|X|X|}
         \hline
         \textbf{Study}  & \textbf{Role Description} & \textbf{User Message} \\
         \hline
         Sexism &  \textbf{Role:} You are a classifier. & \textbf{User:} Classify the following sentence into one of these categories: \newline [sexist, non-sexist].\newline
    Provide your response in the following format: \newline
    Category: [category] \newline
    Explanation: [explanation] \newline
    Sentence: \textit{sentence}:  \\
         \hline
         Topics  & \textbf{Role:} You are a classifier & \textbf{User:} Classify the following sentence into one of these categories: \newline [Economy, Welfare-and Quality of Life, Fabric of Society, Political System, Social Groups, Freedom and Democracy, External Relations].\newline
    Provide your response in the following format: \newline
    Category: [category] \newline
    Explanation: [explanation] \newline
    Sentence: \textit{sentence}:  \\ \\
         \hline
    \end{tabularx}
    \caption{\textbf{Prompts Classification}} % Add a caption here

\end{table*}

\section{Hyperparameters}
\label{hyperpara}

\textbf{RoBERTa Models:}. We use the "roberta-base" model from the \texttt{transformer} library, with 12-layer, 768-hidden, 12-heads, 125M parameters. We use common hyperparameters: batch size of 8, learning rate of 1e-5, and 3 epochs. Unless otherwise noted, we utilize standard hyperparameters from the \texttt{transformer} python library.

\textbf{SVM:} We use the base C-Support Vector Classification module "SVC" from the \texttt{scikit-learn} library, with a linear kernel and default regularization parameter C =1.0. Unless otherwise noted, we utilize standard hyperparameters from the \texttt{sklearn} python library.

\section{Computational Resources}
\label{comp}

For model training, we used a A100 GPU from Google Colab running Python Python 3.10.12 using Jupyter Notebooks. Fine-tuning RoBERTa-base and classifying our in-domain and OOD test sets for both studies took approximately 6 (political topics study) and 8 (sexism study) hours. 

\section{Performance on OOD Set}
\label{perfo_ood}
\begin{figure*}[ht!]
    \centering
    \includegraphics[width=1\linewidth]{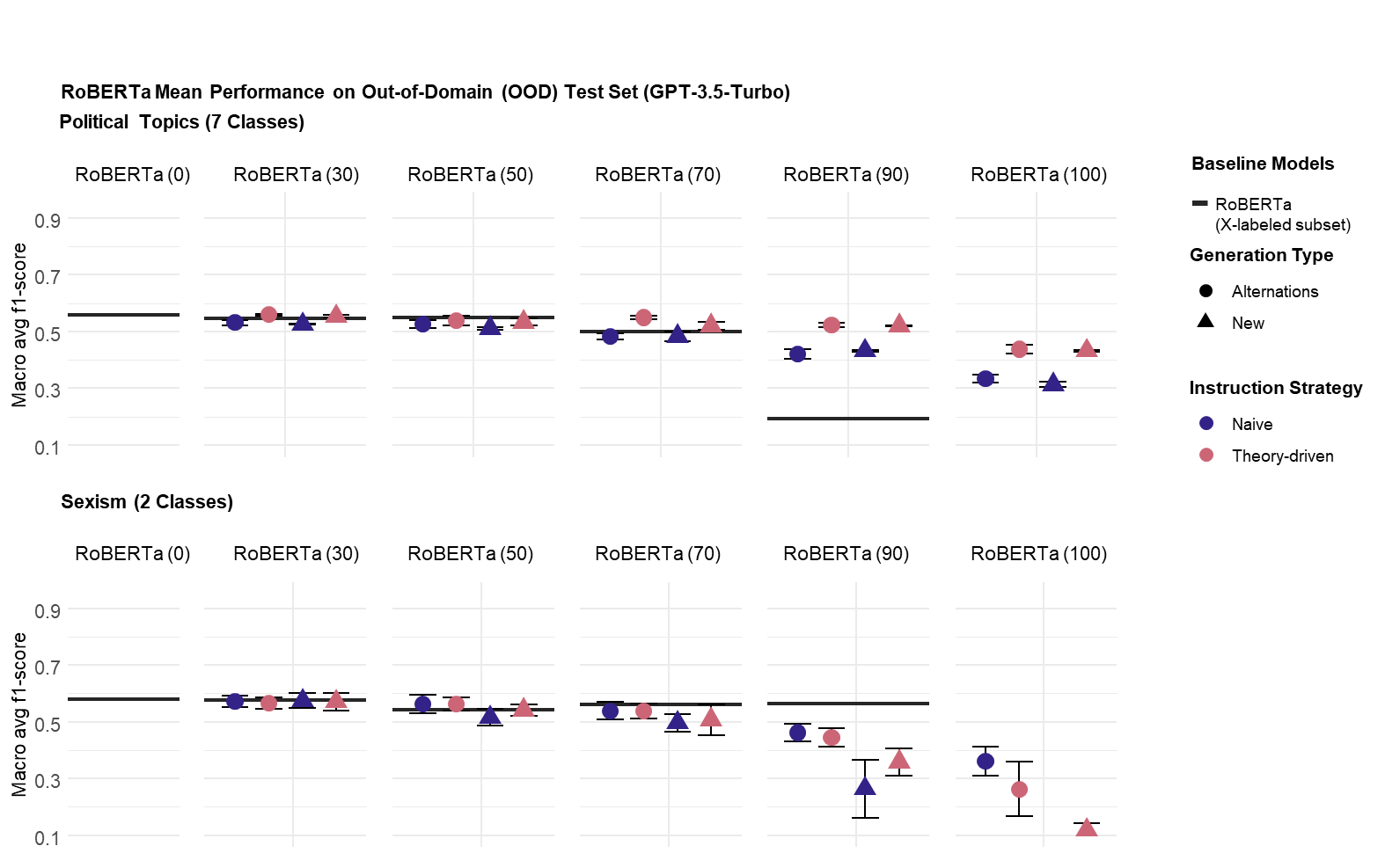}
    \caption{\textbf{Performance on OOD Set}. The y-axis depicts the macro-F1 score across three random seeds. Error bars represent 95\% confidence intervals across the three training runs. Training sizes were balanced across classes, with $N_{Training}$ = 1,938 for the sexism study (969 cases per class) and $N_{Training}$ = 3,500 for the topics study (500 cases per class). The dark grey lines correspond to the results of a model trained on only the labeled subset of the data; synthetic data was generated using GPT-3.5-turbo.}
    \label{fig:ood}
\end{figure*}

\section{Robustness Checks}
\label{robstuness_checks}
\begin{figure*}[h]
    \centering
    \includegraphics[width=0.8\linewidth]{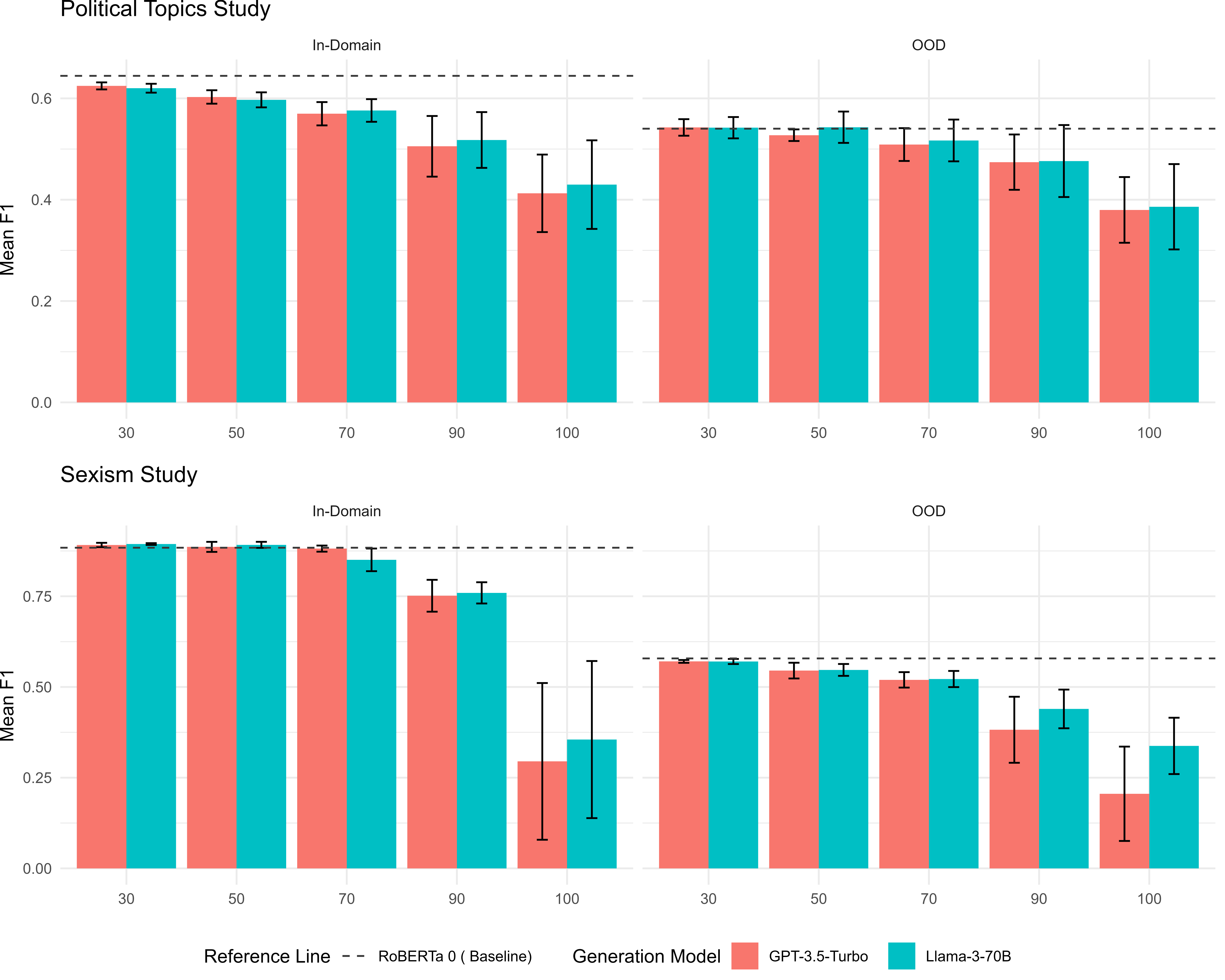}
        \caption{\textbf{Performance and Model Generation}.The y-axis corresponds to the mean performance across data generation strategies and random seeds for each share of synthetic data and generation model. The dark grey lines correspond to the baseline model that was only trained on labeled data.}
    \label{fig:res_3}    
\end{figure*}

To test the robustness of our findings, Figure \ref{fig:res_3} displays the differences in performance between the default GPT-3.5-turbo and the Llama-3-70b generation model. 

In our findings, the performance differences between models generated by Llama-3-70b and GPT-3.5-turbo were minimal, with both models performing similarly across various synthetic data proportions. However, Llama-3-70b showed a slight advantage in out-of-domain (OOD) classification, where its F1-scores were marginally higher than those of GPT-3.5-turbo, particularly as the proportion of synthetic data increased. Despite this edge in OOD scenarios, the overall performance trend remained consistent across both models, indicating that the choice of generation model—whether Llama or GPT—did not significantly impact results, especially in in-domain settings.

\end{appendices}

\end{document}